\documentclass[a4paper,fleqn]{cas-sc}

\usepackage[numbers]{natbib}

\usepackage{subcaption}
\usepackage{orcidlink}
\usepackage{multirow}
\usepackage{adjustbox}
\usepackage{makecell}
\usepackage{placeins}

\def\tsc#1{\csdef{#1}{\textsc{\lowercase{#1}}\xspace}}
\tsc{WGM}
\tsc{QE}

\begin{document}
\let\WriteBookmarks\relax
\def\floatpagepagefraction{1}
\def\textpagefraction{.001}

\shorttitle{Bridging the gap in FER: addressing age bias in deep learning}    

\shortauthors{Gaya-Morey et al.}  

\title [mode = title]{Bridging the gap in FER: addressing age bias in deep learning}  



%
\author[1]{F. Xavier Gaya-Morey}[type=editor,
      auid=000,
      bioid=1,
      orcid=0000-0003-1231-7235]
\cormark[1]
\ead{francesc-xavier.gaya@uib.es}
\ead[url]{https://personal.uib.eu/francesc-xavier.gaya}
\credit{Conceptualization, Methodology, Validation, Investigation, Writing - Original Draft, Writing - Review \& Editing Preparation, Software, Visualization}

\author[1]{Julia Sanchez-Perez}[type=editor,
      auid=001,
      bioid=2,
      orcid=0009-0008-1274-031X]
\ead{julia.sanchez@uib.es}
\credit{Conceptualization, Methodology, Validation, Investigation, Writing - Original Draft, Writing - Review \& Editing Preparation, Software, Visualization}

\author[1]{Cristina Manresa-Yee}[type=editor,
      auid=002,
      bioid=3,
      orcid=0000-0002-8482-7552]
\ead{cristina.manresa@uib.es}
\ead[url]{https://personal.uib.eu/cristina.manresa}
\credit{Conceptualization, Methodology, Writing - Review \& Editing Preparation, Supervision, Project administration, Funding acquisition}

\author[1]{Jose M. Buades-Rubio}[type=editor,
      auid=003,
      bioid=4,
      orcid=0000-0002-6137-9558]
\ead{josemaria.buades@uib.es}
\ead[url]{https://personal.uib.eu/josemaria.buades}
\credit{Conceptualization, Methodology, Writing - Review \& Editing Preparation, Supervision, Project administration, Funding acquisition}




\affiliation[1]{organization={Universitat de les Illes Balears (UIB)},
            addressline={Carretera de Valldemossa, km 7.5}, 
            city={Palma},
            postcode={07122}, 
            state={Illes Balears},
            country={Spain}}

\cortext[1]{Corresponding author}



\begin{abstract}
    Facial Expression Recognition (FER) systems based on deep learning have achieved impressive performance in recent years. However, these models often exhibit demographic biases, particularly with respect to age, which can compromise their fairness and reliability. In this work, we present a comprehensive study of age-related bias in deep FER models, with a particular focus on the elderly population. We first investigate whether recognition performance varies across age groups, which expressions are most affected, and whether model attention differs depending on age. Using Explainable AI (XAI) techniques, we identify systematic disparities in expression recognition and attention patterns, especially for ``neutral'', ``sadness'', and ``anger'' in elderly individuals. Based on these findings, we propose and evaluate three bias mitigation strategies: Multi-task Learning, Multi-modal Input, and Age-weighted Loss. Our models are trained on a large-scale dataset, AffectNet, with automatically estimated age labels and validated on balanced benchmark datasets that include underrepresented age groups. Results show consistent improvements in recognition accuracy for elderly individuals, particularly for the most error-prone expressions. Saliency heatmap analysis reveals that models trained with age-aware strategies attend to more relevant facial regions for each age group, helping to explain the observed improvements. These findings suggest that age-related bias in FER can be effectively mitigated using simple training modifications, and that even approximate demographic labels can be valuable for promoting fairness in large-scale affective computing systems.
\end{abstract}




\maketitle

\section{Introduction}

    Facial expressions play a key role in conveying emotions and intentions in human communication~\cite{darwin1998expression}. Because of their relevance in a wide range of applications---such as assistive robotics, healthcare, driver monitoring, or other human-computer interaction systems---automatic facial expression recognition (FER) has become a prominent research area in computer vision. Over the years, many FER systems have been developed to extract emotional information from facial features, with recent progress being driven largely by deep learning techniques~\cite{li2022deep}.

    Thanks to the unprecedented advances in the machine learning field, many techniques for tackling this task now use deep learning approaches \cite{li2022deep}. These data-driven approaches rely heavily on large annotated datasets, which can inadvertently reflect imbalances present in the collected data~\cite{drozdowski2020demographic}. One of the most common and impactful biases in FER datasets is related to age: children and elderly individuals tend to be underrepresented, while adults are typically overrepresented. Moreover, the facial morphology and expressive patterns of older adults differ from those of younger individuals, as highlighted in prior studies~\cite{raghebatallah2019review}. For example, wrinkles and skin folds can interfere with expression decoding and introduce challenges for both human and automated systems~\cite{fölster2014facial}. Without proper handling of these differences, FER models risk learning spurious correlations and producing inaccurate predictions for minority age groups \cite{suresh2021framework, mehrabi2021survey}.

    To better understand and mitigate such issues, researchers have increasingly turned to eXplainable Artificial Intelligence (XAI)~\cite{arrieta2020explainable, Adadi}, which provides tools to interpret the internal decision-making processes of deep learning models. By offering insights into the features and regions that guide model predictions~\cite{gunning2019xai—explainable}, XAI can be used to detect and characterize biases, as shown in studies examining gender bias~\cite{ijimai} and differences between neurotypical users and those with intellectual disabilities~\cite{ramis_guarinos2024explainable}. However, to date, little attention has been given to leveraging XAI to explore age-related bias in FER.

    In this work, we aim to address this gap by conducting a comprehensive analysis of age-related bias in deep learning models for FER. Using XAI techniques, we examine how such biases manifest across age groups, which expressions are most affected, and how model attention differs depending on the subject’s age. Based on these insights, we implement and evaluate three bias mitigation strategies: Multi-task Learning, Multi-modal Input, and Age-weighted Loss. These approaches are tested on AffectNet, using automatically estimated age labels for training, and validated on age-balanced datasets such as FACES~\cite{ebner2010faces}, RaFD~\cite{langner2010presentation}, and DDCF~\cite{dalrymple2013dartmouth}.


    The main contributions of this work are as follows:
    
    \begin{itemize}
    
        \item We perform an in-depth analysis of age-related biases in deep FER models using XAI tools, identifying how expression recognition and model attention differ across age groups.
        
        \item We propose and evaluate three bias mitigation strategies---Multi-task Learning, Multi-modal Input, and Age-weighted Loss---and show that they improve performance, particularly for elderly individuals.
        
        \item We demonstrate that models trained on AffectNet with automatically estimated age labels can generalize well to other datasets, outperforming previous methods such as Huang et al.~\cite{huang2024facial} on the elderly group from FACES.
        
        \item We show, through heatmap analysis, that age-aware training strategies lead models to focus on more relevant facial regions for each age group, helping to explain the observed improvements in accuracy.
        
    \end{itemize}

    All code is made publicly available at \url{https://github.com/Xavi3398/FER-for-elderly}.
    
    This paper is structured as follows. First, we review related work, focusing on how age influences FER and existing approaches for mitigating age-related bias. We then formulate our research questions, followed by a detailed description of our experimental setup, including models, datasets, XAI tools, and debiasing strategies. Finally, we present our results, discuss key findings, and outline directions for future research.
    

\section{Related work}




        Many psychological studies have investigated the impact of aging on facial expression recognition, revealing that the individuals perceiving the expressions are affected by age-related changes. For instance, older adults often experience difficulties in accurately recognizing certain emotions \citep{isaacowitz2011bringing, ruffman2008meta}. However, beyond perception, the structural changes in aging faces also affect how expressions are displayed and interpreted. Fölster et al. \cite{fölster2014facial} analyzed how age-related facial changes, such as wrinkles and folds, impact the recognition of emotional expressions. Their findings emphasized that facial aging significantly alters the way emotions are conveyed and perceived. This conclusion is reinforced by Ko et al. \cite{ko2021changes}, who observed that older adults not only display more negative emotions but also engage more lower-face muscles than younger individuals. These differences in facial muscle activation suggest that age-related changes can lead to variations in expression intensity, making recognition more complex.

        These age biases in expression recognition can also be found in computational approaches. For instance, Grondhuis et al. \cite{ngrondhuis2021having} applied generative adversarial networks to analyze the increasing difficulty of recognizing expressions in older faces. Their study attributed this challenge to the gradual decline in facial muscle function, which affects the clarity and distinctiveness of expressions. Similarly, Sönmez \cite{battinisonmez2019computational} investigated the effect of training FER models on data from different age groups and found that expression recognition was most challenging for elderly faces. Furthermore, Park et al. \cite{park2022facial} demonstrated the importance of addressing age bias in FER. Their study showed that training models on age-specific datasets led to a 22\% increase in accuracy compared to training on datasets that did not account for age differences. These findings emphasize the necessity of developing age-aware FER models to improve recognition accuracy across diverse age groups. A broader review by Atallah et al. \cite{raghebatallah2019review} outlined several key challenges in age-related FER, including the scarcity of datasets containing images of the same individuals at different ages, the difficulty models face in capturing aging patterns, and the significant variation in aging effects across individuals.
        
        Since the influence of age on facial expression recognition has been widely studied, multiple studies have aimed at mitigating age-related biases \citep{gaya-morey2025deep}. Xu et al. \cite{xu2020investigating} compared a baseline FER model with two bias-aware approaches: one incorporating age as an explicit feature, inspired by Guo et al. \cite{guo2010human} and Dwork et al. \cite{dwork2012fairness}, and another using multiple branches to disentangle age from expression recognition, following the work of Alvi et al. \cite{alvi2018turning} and Liu et al. \cite{liu2018exploring}. Al-Garaawi et al. \cite{al_garaawi2022fully} trained age-specific expression models using shape, texture, and appearance features, combining them through a weighted strategy based on estimated age. Their approach, tested on FACES, LifeSpan, and NEMO, demonstrated promising performance across different age groups.
        
        A different approach was followed by Guo et al. \citep{guo2013facial}, who suggested reducing aging effects using facial smoothing techniques that preserve key structural features while removing age-related details. Wu et al. \cite{wu2015enhanced} developed a Bayesian network integrating age information during training. Using the FACES and LifeSpan datasets, their results showed significant biases in cross-group recognition, but incorporating age improved accuracy for most expressions, except happiness. Later, Yang et al. \cite{yang2018joint} introduced a multi-task learning model that jointly predicted age and expressions, leveraging shared representations from convolutional and scatter networks. Their approach outperformed previous single-task models on the same datasets. A more specialized strategy was taken by Huang et al. \cite{huang2024facial}, who integrated an age group classifier within the recognition pipeline. By tailoring features to different age brackets (infants, adolescents, adults, middle-aged, and elderly), their model improved FER performance, particularly for elderly individuals, when tested on RAF-DB, AffectNet, and FACES.
        
        Some studies have focused on assessing age-related biases in commercial FER systems. Kim et al. \cite{kim2021age} benchmarked four commercial systems using the FACES dataset and found that performance was highest for younger users and lowest for older ones, particularly for expressions like anger and neutrality. Similarly, Zhu et al. \cite{zhu2024defining} evaluated FaceReader \citep{noldus} using GAN-generated images across four age groups, confirming strong agreement with human evaluations for most age groups, except for children.
        
        Dataset-specific studies have also been conducted. Jannat et al. \cite{rahatuljannat2021expression} analyzed the EmoReact and ElderReact datasets using a siamese neural network trained on expression pairs. While within-dataset performance was high, cross-dataset results revealed unexpected patterns, with models trained on EmoReact improving performance on ElderReact but not vice versa. Meanwhile, Caroppo et al. \cite{caroppo2020comparison} compared machine learning and deep learning methods across four datasets (FACES, LifeSpan, FER-2013, and CIFE), categorizing individuals into four age groups. They estimated age labels for FER-2013 and CIFE using landmark-based methods, finding that VGG16 combined with a random forest classifier performed best for older adults, though cross-dataset generalization remained a challenge.






        
\section{Research questions}
\label{sec:questions}

    To guide this work, we formulated a set of research questions, presented in Table~\ref{tab:questions}. These questions aim to analyze both the presence of age-related biases in FER systems and the potential of mitigation strategies to address them. The questions are organized into two main blocks: one focused on understanding the bias, and the other on evaluating solutions.

    \begin{table}[h]
        \caption{Primary and secondary research questions}
        \label{tab:questions}
        \adjustbox{width=\textwidth}{
        \begin{tabular}{ll}
            \toprule
            \textbf{ID} & \textbf{Research Question} \\
            \midrule
            \textbf{RQ1} & 
            \textbf{Are there significant differences in model performance across age groups?}\\
            
            RQ1.1 & 
            Which facial expressions show the greatest performance disparities between age groups?\\
    
            RQ1.2 & 
            Do explainable AI methods reveal age-dependent differences in the facial regions the model relies on for FER??\\
            
            \midrule
    
            \textbf{RQ2} & 
            \textbf{Can facial expression recognition models be improved for specific age groups through bias-mitigation techniques?}\\

            RQ2.1 &
            Can the performance gap between facial expressions across age groups be reduced using these techniques? \\
    
            RQ2.2 &
            Do bias-reduction techniques lead to changes in the facial regions used for prediction across age groups?\\
            
            \bottomrule
        \end{tabular}
        }
    \end{table}

    The first group of questions begins with RQ1, which examines whether FER models exhibit noticeable performance differences between age groups. Based on the literature, such age-related disparities have been previously reported; this question aims to confirm and precisely characterize them in the context of trained AI models. To deepen this analysis, RQ1.1 explores whether certain expressions are more affected than others, identifying which emotions are harder to recognize in specific age groups. This allows us to uncover patterns that may relate to how emotional expressions evolve with age. In parallel, RQ1.2 investigates whether the facial regions that models rely on vary depending on the age of the subject. Differences in attention maps across groups may reflect changes in facial features due to aging---such as wrinkles, sagging, or decreased muscle tone---and could reveal how models implicitly adapt (or fail to adapt) to such differences.

    The second block focuses on evaluating mitigation strategies. RQ2 addresses whether recognition performance for specific age groups---particularly underrepresented ones---can be improved by applying bias-reduction techniques. From here, RQ2.1 asks whether these improvements are consistent across expressions, aiming to determine if such techniques not only improve accuracy but also promote a fairer distribution of performance across emotions and age groups. Finally, RQ2.2 complements the analysis of RQ1.2 by examining whether mitigation techniques change the facial regions the models attend to. If such changes are observed, they could indicate that the model has adapted its internal representation to better suit the facial characteristics of different age groups.

\section{Methods}

    \subsection{Deep learning models}
    \label{sec:networks}

        To efficiently run the multiple trainings and evaluations required to address our research questions, we selected three different networks: MobileNetV3~\cite{MobileNetV3}, SwinB~\cite{SwinTransformer}, and ConvNext-Base~\cite{ConvNext}.
        
        MobileNets~\cite{MobileNet} employ depthwise separable convolutions, which significantly reduce the number of parameters and computations compared to traditional convolutions. MobileNetV3, the latest iteration, combines this efficient architecture with platform-aware neural architecture search (NAS) and novel components such as squeeze-and-excitation modules and hard-swish activation functions. With its 5.5M parameters on its large version, MobileNetV3 achieves state-of-the-art performance on benchmarks like ImageNet while maintaining low latency and energy consumption, making it well-suited for multiple training iterations with reduced time cost.
        
        Swin Transformers~\cite{SwinTransformer} introduce hierarchical representation of feature maps and a shifted-window partitioning mechanism between consecutive self-attention layers as key elements of its design, enabling them to achieve linear computational complexity to input image size while mantaining low latency and similar modeling power. The use of these architectural innovations have proved to be effective and efficient on a range of computer vision tasks, including image classification, object detection and semantic segmentation, where Swin Transformers have shown strong performance. Its base variant, SwinB, comprises approximately 88M parameters, thus being substantially more complex than lightweight MobileNets. Consequently, it incurs a higher computational cost during both training and inference phases.
        
        ConvNexts~\cite{ConvNext} arise as an attempt to achieve Vision Transformer (ViTs) performance, while preserving the simplicity and efficiency of standard ConvNets. Its architecture incorporates several enhacements, some inspired by ViTs, including the use of Layer Normalization, the adoption of larger convolutional kernel sizes and inverted bottleneck structures and the reduction of activation functions in residual blocks, among other modifications. Similar to the base Swin Transformer model, ConvNext base variant consists of approximately 88M parameters. It demonstrates competitive, and in some cases superior, performance relative to Swin Transformer across a variety of computer vision benchmarks.

        We implemented these models using the PyTorch framework, initializing with pre-trained weights from ImageNet. The final fully connected layer was replaced to match the number of expression classes (seven in total).

    \subsection{FER datasets}
    \label{sec:datasets}

\begin{figure}[h]
            \captionsetup[subfigure]{justification=centering}
             \centering
             \begin{subfigure}[b]{0.49\textwidth}
                 \centering
                 \includegraphics[height=12em]{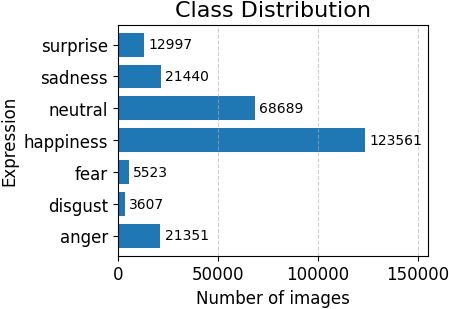}
                 \caption{}
                \label{fig:affectnet-classes}
             \end{subfigure}
             \begin{subfigure}[b]{0.49\textwidth}
                 \centering
                 \includegraphics[height=12em]{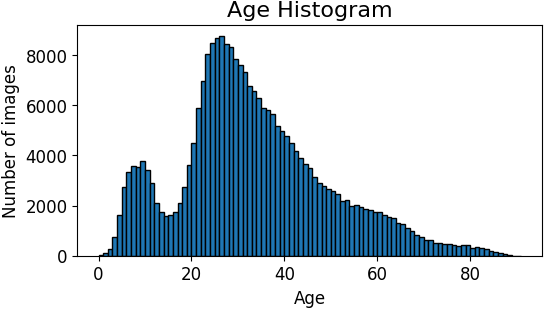}
                 \caption{}
                \label{fig:affectnet-age}
             \end{subfigure}
            \caption{(a) Class distribution and (b) age histogram of the AffectNet dataset.}
            \label{fig:affectnet}
        \end{figure}

        As the training dataset, we used AffectNet~\cite{mollahosseini2017affectnet}, one of the largest publicly available FER datasets, which offers high variability in both image content and user demographics, though it lacks demographic annotations. To enable age-related experiments, we enriched AffectNet with age labels using the MiVOLO network~\cite{kuprashevich2024mivolo}, a transformer-based model designed for age and gender estimation. MiVOLO achieves a mean absolute error (MAE) of around 5 on popular age estimation datasets like IMDB~\cite{rothe2015dex} and UTKFace~\cite{zhang2017age}. While automatic age labeling introduces some noise compared to ground truth age annotations, it enabled fast and straightforward age-aware experiments that would otherwise be impossible on AffectNet. We used the duplicate-free version of the dataset~\cite{ijimai}, removing all images labeled as ``contempt,'' as well as images lacking annotations, a detectable face, or with uncertain labels. This filtering yielded a total of 257,168 images. Figure~\ref{fig:affectnet} shows the class and age distributions.

        To evaluate model performance across age groups with reduced evaluation bias \cite{suresh2021framework}, we compiled several datasets to represent each age group based on two criteria: (i) they must be static (image-based), and (ii) recorded in controlled environments. We excluded datasets composed of video sequences, frames extracted from videos, or scraped from the Internet. This ensured that each image sample was reliably labeled and free from low-intensity or transitional expressions. We also excluded all non-frontal images. Based on these criteria, the selected datasets per age group were:

        \begin{itemize}
        
            \item Children (-18): NIMH-ChEFS~\cite{egger2011nimh}, RaFD~\cite{langner2010presentation}, DDCF~\cite{dalrymple2013dartmouth}, and DEFSS~\cite{meuwissen2017creation}
            
            \item Adults (18-59): FEGA~\cite{ramis2022novel}, RaFD~\cite{langner2010presentation}, KDEF~\cite{d_1998karolinska}, WSEFEP~\cite{olszanowski2014warsaw}, and FACES~\cite{ebner2010faces}
            
            \item Elderly (60+): FACES~\cite{ebner2010faces}
            
        \end{itemize}

        While we collected four and five datasets for the children and adult groups, respectively, only one dataset (FACES) contained elderly individuals. Although the Lifespan dataset~\cite{minear2004lifespan} includes some elderly faces, it was excluded due to class imbalance and a very limited number of elderly samples per class (e.g., only 12 for anger, 10 for surprise, 9 for sadness, and none for disgust or fear).

\begin{figure}[h]
            \captionsetup[subfigure]{justification=centering}
             \centering
             \begin{subfigure}[b]{0.335\textwidth}
                 \centering
                 \includegraphics[width=\textwidth]{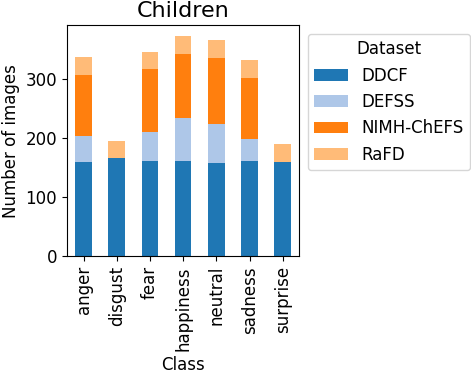}
                 \caption{}
                \label{fig:image-count-children}
             \end{subfigure}
             \begin{subfigure}[b]{0.31\textwidth}
                 \centering
                 \includegraphics[width=\textwidth]{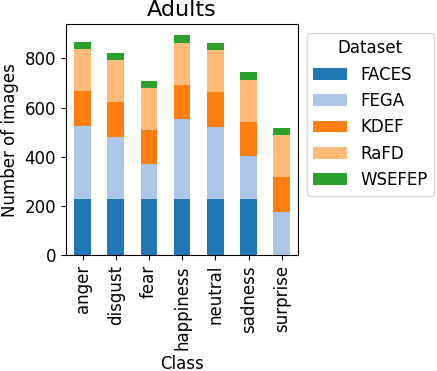}
                 \caption{}
                \label{fig:image-count-adults}
             \end{subfigure}
             \begin{subfigure}[b]{0.3\textwidth}
                 \centering
                 \includegraphics[width=\textwidth]{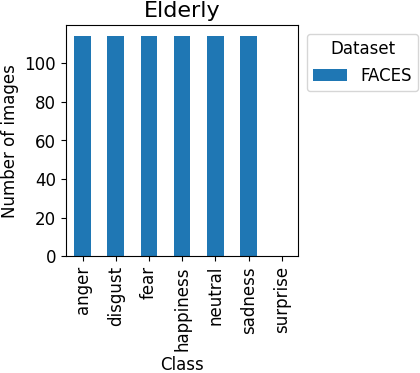}
                 \caption{}
                \label{fig:image-count-elderly}
             \end{subfigure}
            \caption{Number of test images per age group: (a) children, (b) adults, and (c) elderly.}
            \label{fig:image-count}
        \end{figure}

        Figure~\ref{fig:image-count} shows the number of images per expression, dataset, and age group. We included the six basic expressions (happiness, sadness, surprise, fear, anger, and disgust), as defined by Ekman~\cite{ekmanuniversal}, plus the neutral expression. However, not all expressions were present in all datasets. For instance, DEFSS and NIMH-ChEFS lack ``disgust'' and ``surprise,'' while FACES lacks ``surprise.'' As seen in the figure, the image count per expression is approximately: 200+ for children, 500+ for adults, and exactly 114 for elderly (with none for ``surprise'').

    \subsection{XAI tools}        

        To better understand the network’s predictions, we leveraged explainable AI tools. Specifically, since we were interested in identifying the most relevant facial regions for each target expression, we adopted the method proposed by Ramis et al.~\cite{ijimai}---originally designed to explore gender bias in FER---and later refined by Gaya-Morey~\cite{gaya_morey2024unveiling} to analyze the similarity between human and machine perception in FER. This method comprises three main steps: (i) computation of local explanations (i.e., per-image explanations); (ii) normalization of the explanations to a common space; and (iii) aggregation into per-expression heatmaps by averaging the standardized explanations.

        Although the original approach employs LIME~\cite{ribeiro2016why} to generate local explanations---requiring multiple forward passes per image (typically over 1,000)---we replaced it with saliency maps~\cite{simonyan2014deep}, which only require a single network pass. Beyond the clear computational advantage, saliency maps---being gradient-based---typically yield higher-fidelity explanations than LIME, as demonstrated by Miro et al.~\cite{MIRONICOLAU2024104179}.
        
        The final output consists of one heatmap per expression, summarizing the five cross-validation training iterations. These heatmaps highlight the most influential regions the network uses to predict each expression. All explanations were computed on the test datasets described in Section~\ref{sec:datasets}.

    \subsection{Training}
    \label{sec:training}

        Before training, we preprocessed the AffectNet dataset following several steps: (i) face detection using YOLOv8~\cite{jocher2023ultralytics}, cropping the image to the detected face; (ii) rotation to align the eyes horizontally, using SPIGA for facial landmark detection~\cite{prados_torreblanca2022shape}; (iii) grayscale conversion; (iv) contrast stretching; and (v) resizing to \(224 \times 224\) pixels. Data augmentation was also applied during training, including random horizontal flipping, rotation, translation, zoom, brightness, and contrast adjustments.
        
        For each of the networks described in Section~\ref{sec:networks}, we performed five training iterations using cross-validation on AffectNet. Each network was trained for up to 20 epochs, with early stopping if validation accuracy did not improve by at least 1\% over five consecutive epochs. We used the Adam optimizer with a learning rate of \(10^{-4}\) and a batch size of 64. Additionally, due to class imbalance in AffectNet (see Figure~\ref{fig:affectnet-classes}), we applied class-weighted cross-entropy loss. This allowed to mitigate learning bias toward majority classes \cite{suresh2021framework}.

    \subsection{Approaches facing age biases}
    \label{sec:approaches}

        We implemented three strategies to reduce age-related bias: Age-weighted Loss, Multi-task Learning, and Multi-modal Input. Since the weighted loss approach is compatible with the others and showed improved performance when combined, we applied age-based weighting in both the multi-task and multi-modal setups. This allowed those approaches to not only incorporate age information but also emphasize minority age groups. Each approach is detailed below.

        \subsubsection{Age-weighted Loss}
        \label{sec:age-weighted}
        
            The simplest mitigation strategy involved applying class weights to both the facial expression and the age group, without modifying the network architecture. Let $\mathcal{S}$ denote the set of facial expression classes and $\mathcal{A}$ the set of age groups. For each expression class $s \in \mathcal{S}$ and age group $a \in \mathcal{A}$, let $N_{s,a}$ denote the number of training samples belonging to expression $s$ and age group $a$. Let $\displaystyle N_{\min} = \min_{s' \in \mathcal{S}, a' \in \mathcal{A}} N_{s',a'}$ represent the smallest class cardinality for expressions and age groups.

            Then, the total weight $w_{s,a}$ assigned to a training sample belonging to expression $s$ and age group $a$ is given by:
            
            \begin{equation}
                w_{s,a} = \frac{N_{\min}}{N_{s,a}}
                \label{eq:weights}
            \end{equation}
            
            This weighting scheme increases the contribution of underrepresented classes (both in terms of expression and age group) during training, encouraging the models to balance their performance across the entire population.

        \subsubsection{Multi-task Learning}
        \label{sec:multitask}

            In the multi-task setup, we hypothesized that a subject’s age is relevant to expression recognition. Therefore, we trained the models to predict both expression and age. By forcing the networks to also estimate age (as a continuous value), they were encouraged to learn age-related features, which are stored in the shared feature vector used for both tasks.

            To enable this, we added an additional linear layer to each architecture, parallel to the one used for expression classification. We used mean squared error (MSE) loss for the age regression task and combined it with the cross-entropy loss from the expression classification task using a weighted sum. Moreover, since AffectNet is imbalanced with respect to age distribution (see Figure~\ref{fig:affectnet-age}), we applied density-based weighting~\cite{steininger2021density-based} to the MSE loss, a technique well-suited for imbalanced regression.
        
        \subsubsection{Multi-modal Input}
        \label{sec:multi-modal-input}

            In this approach, we incorporated the subject’s age as an explicit input to the expression classifier, creating a multi-modal setup. The rationale is to make the networks aware of potential age-related biases and adapt their predictions accordingly.

            To integrate age information, we followed a variant of the attribute-aware approach~\cite{xu2020investigating}. While the original method upsamples the attribute (e.g., age) to match the size of the feature vector via a fully connected layer and then adds the vectors, we instead chose to downsample the visual feature vector via a fully connected layer and concatenate it with the age scalar. This concatenated vector is then passed to a final fully connected layer to predict the expression.
    
    \subsection{Procedure}

        The experimental procedure consisted of three main stages: training, evaluation, and analysis using XAI tools. We trained MobileNetV3, Swin Transformer and ConvNext models in four configurations: the baseline (without age bias mitigation) and three versions corresponding to the methods described in the previous section. Each configuration was trained using the protocol from Section~\ref{sec:training}, resulting in 3 (networks) $\times$ 4 (methods) $\times$ 5 (cross-validation folds) = 60 training runs.

        Second, we evaluated each model by age group and expression class using the test datasets described in Section~\ref{sec:datasets}, averaging the results across the five folds. We used the F1 score as the primary metric, as it is more appropriate than accuracy for imbalanced datasets~\cite{branco2016survey}. We further analyzed performance by examining confusion matrices and per-class F1 scores.
        
        Finally, we applied XAI tools to examine differences between age groups. We performed this analysis first for the baseline model and then for the best-performing age-bias mitigation method. This allowed us to identify and compare the facial regions deemed most relevant by biased versus unbiased versions of the network.

\section{Results}

    \subsection{Age bias exploration}
    \label{sec:age-bias-exploration}

\begin{figure}[h]
            \centering
            \includegraphics[width=\textwidth]{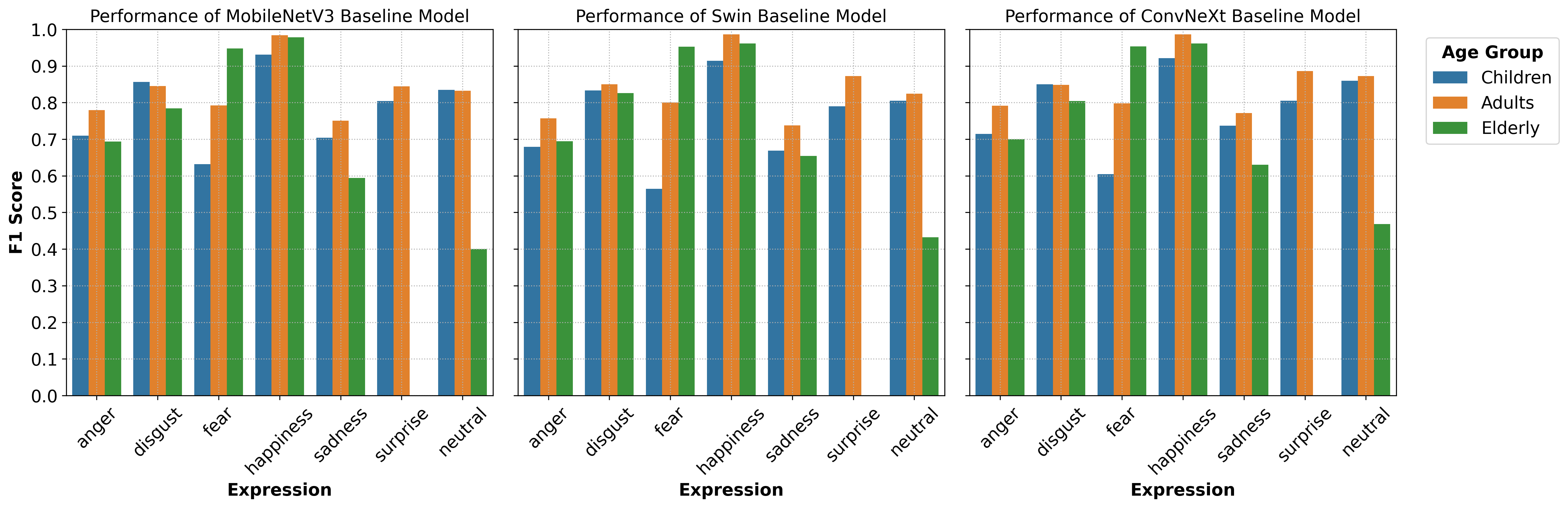}
            \caption{Performance of the baseline by expression and age group. From left to right: MobileNetV3, Swin Transformer and ConvNeXt-Base. Note that the ``surprise'' expression is missing in the elderly group. }
            \label{fig:bars-base}
        \end{figure}

\begin{figure}[ht]
            \centering
            \includegraphics[width=\linewidth]{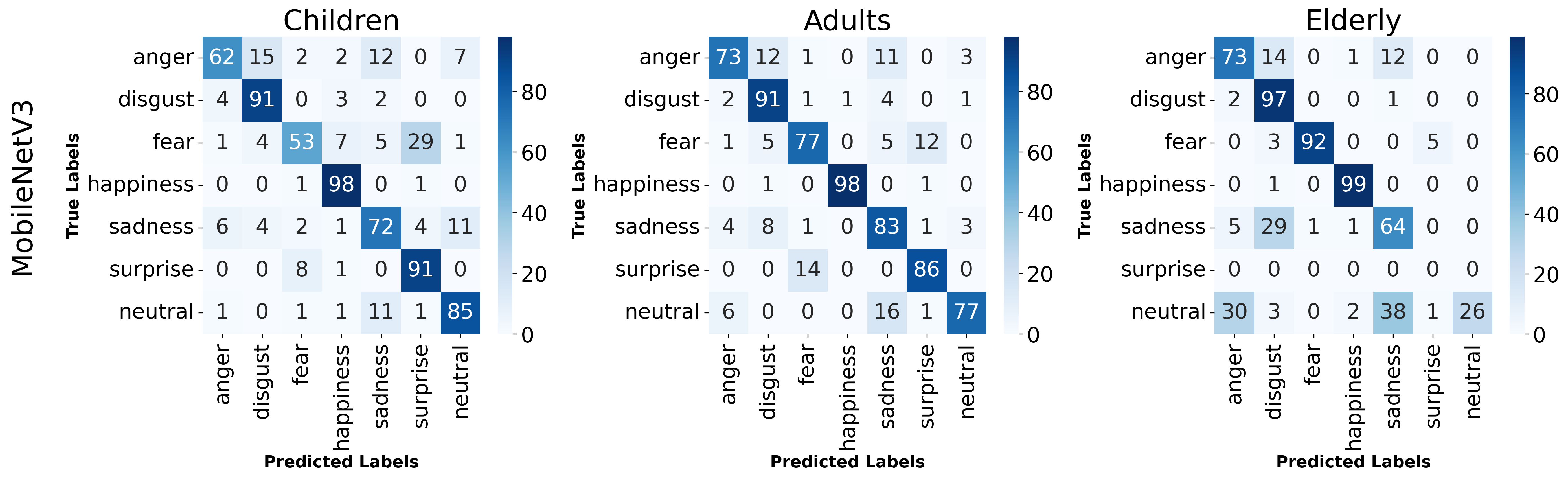} \\[1ex]
            \includegraphics[width=\linewidth]{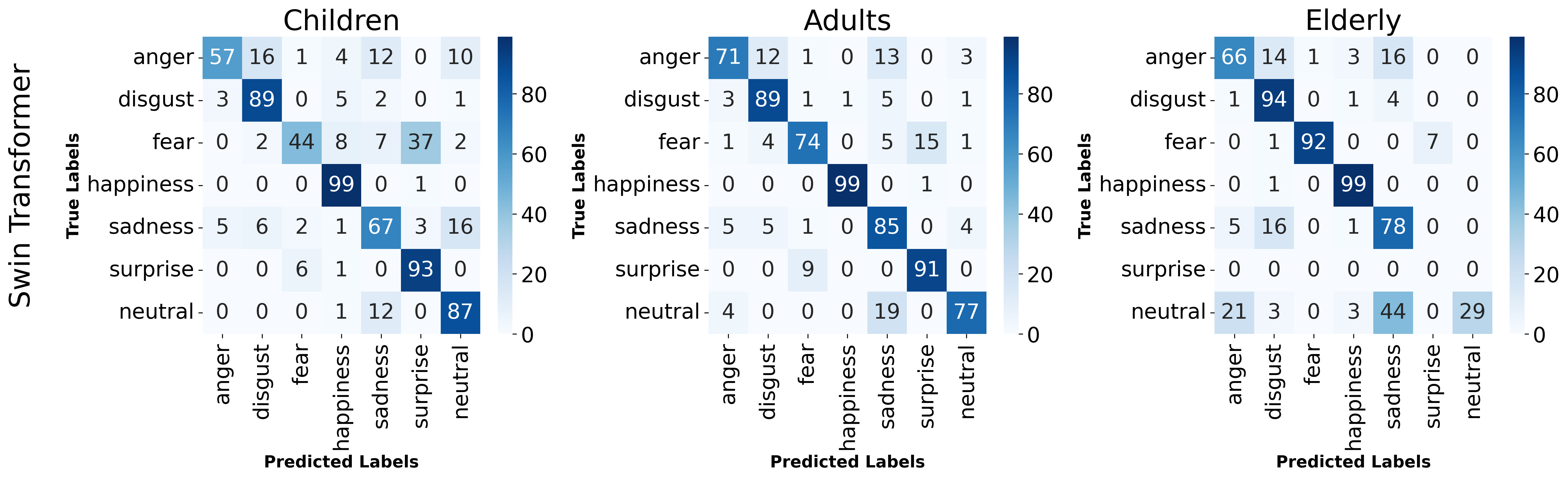} \\[1ex]
            \includegraphics[width=\linewidth]{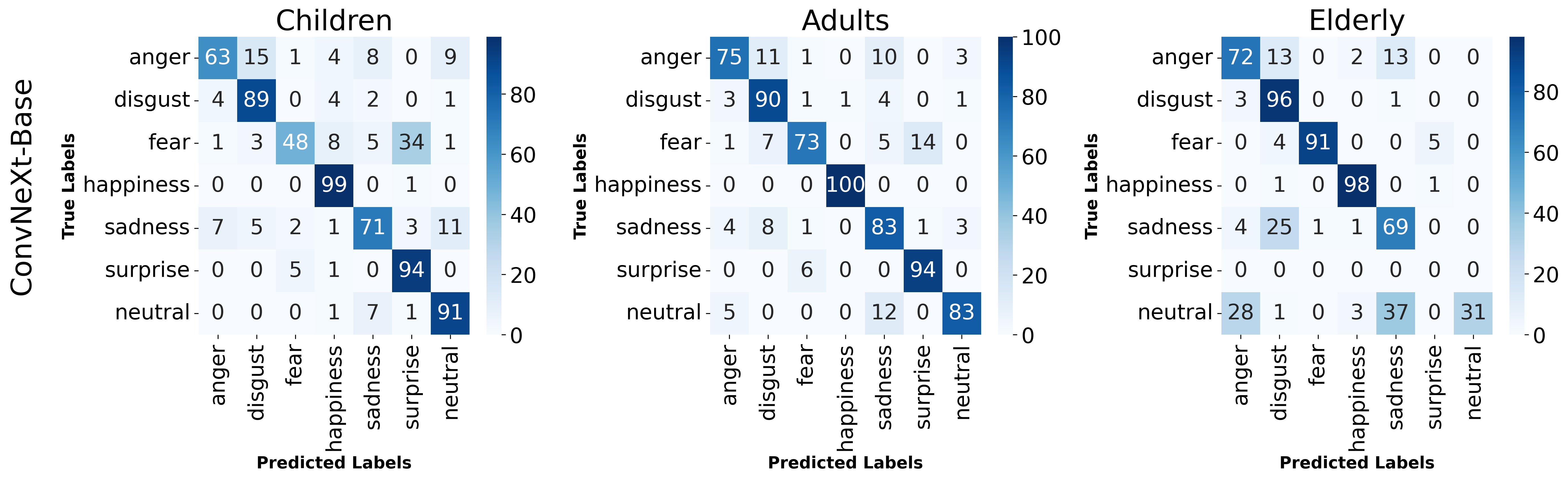}
            \caption{Normalized confusion matrices by age group for the baseline approach. From left to right: children, adults, and elderly. From up to bottom: MobileNetV3, Swin Transformer and ConvNeXt-Base.}
            \label{fig:cms-base}
        \end{figure}

        To explore age bias in the baseline networks, we computed the F1 score by expression and age group, shown in Figure~\ref{fig:bars-base}. We also computed confusion matrices separately for each age group to better understand which expressions were misclassified (Figure~\ref{fig:cms-base}). Since the number of test images varies across expressions, we normalized the confusion matrices by dividing the values of each row by the total number of samples in that row. The resulting values can be interpreted as the percentage of true samples from a given class that were predicted as another given class. This normalization enables fairer comparisons across both expressions and age groups.

        As expected, we observed clear differences between age groups for most expressions. The observations and trends detailed below are consistently identified in the three networks analyzed (MobileNetV3, Swin Transformer and ConvNeXt-Base), indicating coherence in the behavior of the models and suggesting robustness in the patterns identified across the different categories and age groups, regardless of the architecture employed.
        
        The largest difference was found for the ``neutral'' expression, where F1 scores for children and adults remain consistent and comparable to those observed for other expressions, but drop drastically for the elderly. This severe performance drop is also visible in the elderly confusion matrices across all three networks analyzed (Figure~\ref{fig:cms-base}). Interestingly, these confusions also appear in the adult group, albeit to a lesser extent. Conversely, in the children group, ``sadness'' and ``anger'' are frequently misclassified as ``neutral''.

        Another expression with notable age-related differences is ``fear'', for which F1 scores increase with age. This improvement is mainly due to reduced confusion with the ``surprise'' expression.

        For the ``sadness'' expression, performance also drops in the elderly group. This decrease is primarily due to confusion with the ``disgust'' expression, a confusion considerably less frequent in the other groups.

        In contrast, performance for ``anger'', ``disgust'', ``happiness'', and ``surprise'' varies by less than .1 across age groups, with no major differences in the confusion matrices beyond those previously discussed. Note that no ``surprise'' samples are available for the elderly group, as this class is not present in the FACES dataset.

\begin{figure}[h]
            \centering
            \includegraphics[width=\textwidth]{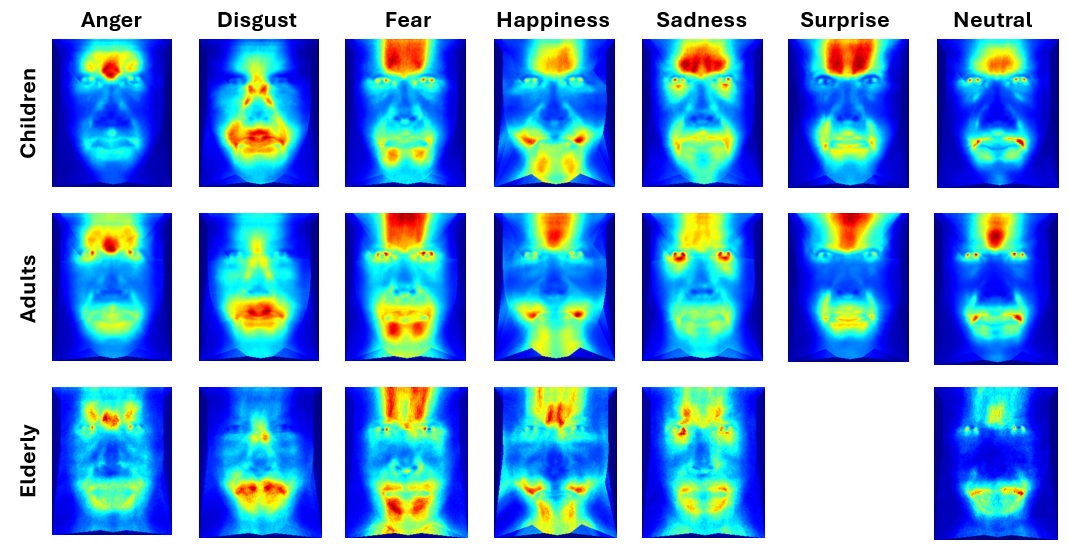}
            \caption{Explanation heatmaps by age group and expression for the ConvNeXt-Base baseline approach. }
            \label{fig:heatmaps-base}
        \end{figure}

        Additionally, we computed explanation heatmaps to explore which facial regions were considered most relevant by the model. Figure~\ref{fig:heatmaps-base} shows explanation heatmaps for the ConvNeXt-Base baseline approach. Overall, we observed considerable variation in the salient regions across expressions, but relatively little variation across age groups.

        Among the expressions, ``happiness'' and ``disgust'' exhibited the most distinct activation patterns. For ``happiness'', the model consistently focused its attention on regions such as the forehead and the edges of the mouth, while for ``disgust'' it emphasized the nose and the entire area of the mouth, focusing on the lips rather than just the contours of the mouth. These clearly defined distinctive activation patterns may help explain the high classification performance for these expressions across all age groups.

        In contrast, the heatmaps for ``sadness'', ``anger'', and ``neutral'' highlighted overlapping regions---primarily the forehead, the corners of the lips, and the eyes. This similarity in saliency may underlie the confusion frequently observed among these expressions, particularly in the elderly group. This resemblance is even more pronounced in the saliency heatmaps for ``fear'' and ``surprise,'' especially among children, which may once again account for the persistent misclassification observed between these expressions.

        Although heatmaps across age groups are broadly consistent, the greatest differences were observed in the elderly group. For example, in the case of ``disgust,'' the nose appears to play a less important role in the elderly group compared to the others. Likewise,  for ``sadness'' the salience of the forehead and frown regions appears to diminish progressively with increasing age, as it happens with ``neutral'' expression too.

        Heatmaps for the Swin Transformer and MobileNetV3 models for the baseline approach were also computed. A comparative analysis on the three of them is presented on Appendix~\ref{secA1}.

    \subsection{Improvements for minority groups}

\begin{figure}[h]
            \centering
            \includegraphics[width=\textwidth]{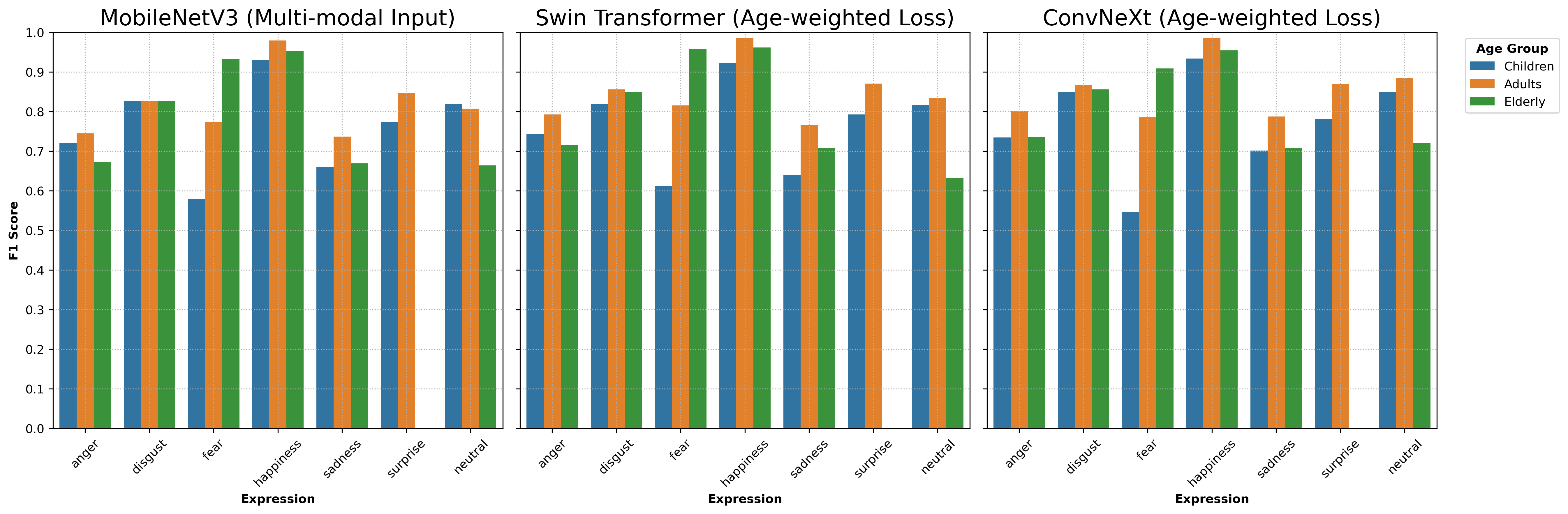}
            \caption{Performance of the best approach based on macro average F1 score by expression and age group. From left to right: MobileNetV3 Multi-modal Input, Swin Transformer Age-weighted Loss and ConvNeXt-Base Age-weighted Loss.}
            \label{fig:bars-best}
        \end{figure}

\begin{figure}[h]
            \centering
            \includegraphics[width=\textwidth]{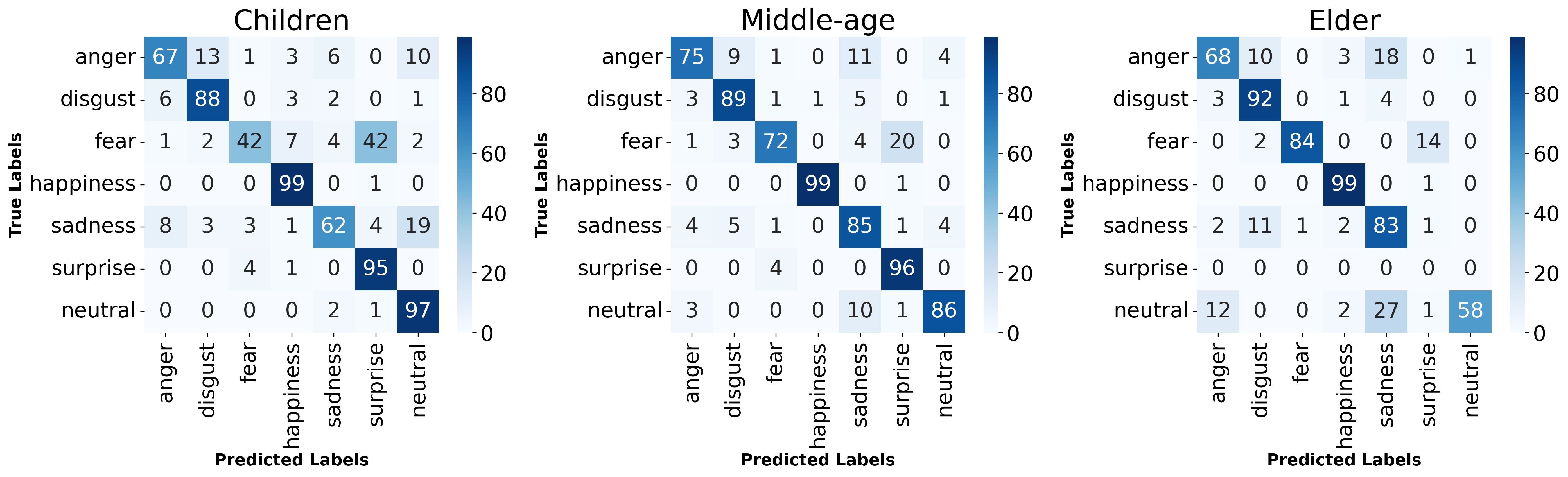}
            \caption{Normalized confusion matrices by age group for the ConvNeXt-Base Age-weighted Loss approach. From left to right: children, adults, and elderly.}
            \label{fig:cms-convnext-age-weighted}
        \end{figure}

        Among the methods designed to reduce age bias, the approach that showed the greatest improvement for elderly participants was the ConvNeXt-Base model with Age-weighted Loss, followed closely by Swin Transformer also with Age-weighted Loss, and finally by the ConvNeXt-Base with the Multi-modal Input approach. In most cases, this enhancement on the elderly performance was accompanied by a slight decrease in performance across the other age groups. The remaining average results can be found in Table~\ref{tab:results}. Additionally, Figure~\ref{fig:bars-best} displays the F1 scores, categorized by expression and age group, for the best-performing approach corresponding to each of the three network architectures based on the average F1 scores reported in Table~\ref{tab:results}, while Figure~\ref{fig:cms-convnext-age-weighted} shows the confusion matrices for the ConvNeXt-Base age-weighted loss approach.
        

        \begin{table}
            \caption{Macro average F1 scores (mean - std) by method and age group (excluding ``surprise''). In bold, the higher score for each column group.}
            \label{tab:results}
            \centering
            \setlength{\tabcolsep}{3pt}
            \begin{tabular}{ll|cc|cc|cc|cc}
                \toprule
                \multirow{2}{*}{\textbf{Method}} & \multirow{2}{*}{\textbf{Network}}
                & \multicolumn{2}{c|}{\textbf{Children}} 
                & \multicolumn{2}{c|}{\textbf{Adults}} 
                & \multicolumn{2}{c|}{\textbf{Elderly}} 
                & \multicolumn{2}{c}{\textbf{Average}} \\
                \cmidrule(lr){3-10}
                & & Mean & Std & Mean & Std & Mean & Std & Mean & Std \\
                \midrule
                Baseline & MobileNetV3     & .7779 & .0172 & .8305 & .0108 & .7329 & .0169 & .7804 & .0150 \\
                Baseline & Swin Transf.    & .7442 & .0062 & .8261 & .0100 & .7535 & .0165 & .7746 & .0109 \\
                Baseline & ConvNeXt        & .7810 & .0162 & .8445 & .0080 & .7528 & .0182 & .7928 & .0141 \\
                \midrule
                Age-weighted Loss & MobileNetV3  & .7565 & .0108 & .8035 & .0142 & .7620 & .0179 & .7740 & .0143 \\
                Age-weighted Loss & Swin Transf. & .7588 & .0132 & .8416 & .0042 & .8043 & .0099 & .8016 & .0091 \\
                Age-weighted Loss & ConvNeXt     & .7695 & .0231 & \textbf{.8520} & .0030 & \textbf{.8140} & .0135 & \textbf{.8118} & .0132 \\
                \midrule
                Multi-task Learning & MobileNetV3  & .7644 & .0185 & .8094 & .0202 & .7799 & .0198 & .7846 & .0195 \\
                Multi-task Learning & Swin Transf. & .7814 & .0199 & .8356 & .0205 & .7509 & .0365 & .7893 & .0257 \\
                Multi-task Learning & ConvNeXt     & \textbf{.7856} & .0070 & .8400 & .0086 & .7935 & .0217 & .8064 & .0124 \\
                \midrule
                Multi-modal Input & MobileNetV3  & .7562 & .0218 & .8115 & .0139 & .7863 & .0241 & .7847 & .0199 \\
                Multi-modal Input & Swin Transf. & .7355 & .0245 & .8227 & .0119 & .7824 & .0247 & .7802 & .0204 \\
                Multi-modal Input & ConvNeXt     & .7850 & .0185 & .8460 & .0131 & .7979 & .0086 & .8096 & .0134 \\
                \bottomrule
            \end{tabular}
        \end{table}

        The greatest improvements were observed in the elderly group. The most significant gain was for the ``neutral'' expression. Comparing the confusion matrices for the baseline (Figure~\ref{fig:cms-base}) and age-weighted loss (Figure~\ref{fig:cms-convnext-age-weighted}) models for ConvNeXt-Base, we observe a clear reduction in misclassifications of ``neutral'' as ``anger'' and ``sadness'', and also of ``sadness'' as ``disgust'', resulting in a substantial increase in true positives. On the other hand, performance for ``fear'', ``disgust'' and ``anger'' decreased slightly for elderly individuals. The confusion matrix suggests that the drop in ``fear'' performance is due to increased confusion with ``surprise'', which is also present in other age groups. For the children and adult groups, differences between methods were smaller. Examples of expressions with greater variation include ``fear'' and ``sadness'' for children. For the remaining expressions, no significant deterioration in prediction performance was observed compared to the baseline. For a more detailed analysis, see Appendix~\ref{secA1}.

\begin{figure}[h]
            \centering
            \includegraphics[width=\textwidth]{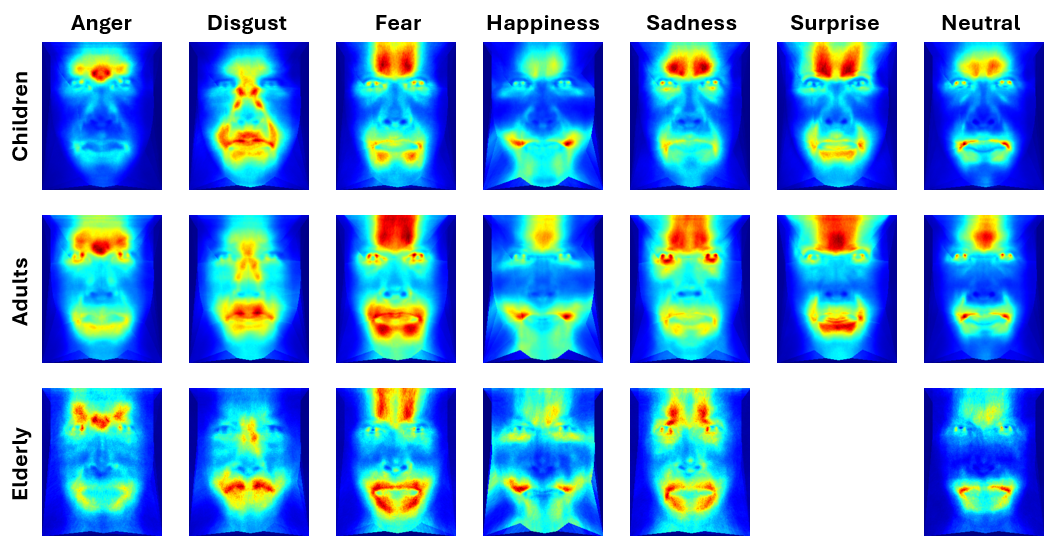}
            \caption{Explanation heatmaps by age group and expression for the ConvNeXt-Base Age-weighted Loss approach.}
            \label{fig:heatmaps-convnext-age-weighted}
        \end{figure}

        To better understand the performance improvements observed in the elderly group, we computed explanation heatmaps for the ConvNeXt-Base Age-weighted Loss model, shown in Figure~\ref{fig:heatmaps-convnext-age-weighted}. When compared to the baseline heatmaps (Figure~\ref{fig:heatmaps-base}), the activation patterns for the ``anger'', ``disgust'', ``fear'' and ``neutral'' expressions remained largely unchanged. However, some differences were found in the remaining expressions. For ``happiness'', the model assigned less relevance to the forehead and frown; for ``sadness'', the model increased focus on the frown and eyebrows regions for adults and elderly individuals, and gave even more importance to the edges of the mouth for the elderly group; and for ``surprise'', the model paid more attention to the mouth area, while maintaining its attention on the forehead region.
        
\section{Discussion}

    To structure our discussion, we revisit the two research questions outlined in Section~\ref{sec:questions}.

    \subsection{RQ1: Are there significant differences in model performance across age groups?}

        According to the obtained results, it is clear that the baseline model is not free of age bias. The results obtained from the baseline exploration phase, consisting of per-age-group and per-expression performance, confusion matrices, and heatmaps, revealed three main biases related to age: (i) especially high confusion of neutral with sadness and anger for the elderly; (ii) especially high confusion of fear with surprise for children; and (iii) high confusion of sadness with disgust, observed only in the elderly. Evidently, these confusion problems hindered model performance for the affected expressions and age groups, with the elderly being the most affected.

        The F1 score for the neutral expression appears to follow an inverse relation with age: the higher the age of the subjects, the more a neutral expression is predicted as sadness or anger. In fact, for children, sadness and anger expressions are mistakenly predicted as neutral---the opposite case. This phenomenon is likely due to age-related facial differences, such as wrinkles and folds, compounded by the similarity in the heatmaps of these expressions (see Figure~\ref{fig:heatmaps-base}). Since the model focuses its attention on roughly the same facial regions for the three expressions, age-related changes in these regions may confuse the model, particularly if minority groups are underrepresented in the training data, which is the case of the dataset used for training---AffectNet.

        We observed a positive relation between age and the F1 score for the ``fear'' expression, which is less confused with ``surprise'' as age increases. Heatmap inspection revealed the importance of the forehead in predicting ``surprise'' and ``fear''. A plausible hypothesis is that the model learns to associate expression wrinkles in the forehead---related to action units like AU1 (inner brow raiser), AU2 (outer brow raiser), and AU5 (upper lid raiser)~\cite{ekman1978facial}---with ``fear''. Therefore, the absence of such wrinkles in children may confuse the model, leading it to underpredict ``surprise''. Interestingly, the confusion between fear and surprise in children has been widely reported in psychology literature~\cite{bullock1985further, gosselin1995le, gosselin1999childrens}. Hence, a possible explanation is that the difficulty children have in distinguishing between these expressions is reflected in how they express them, diverging from the patterns observed in other age groups. Different from the other observed bias, this can pose a limit in supervised deep learning models performance, highly dependent on subjective annotations.

        The third main bias was the lower performance on ``sadness'' for elderly individuals, which was often confused with ``disgust''. Unlike ``sadness'', the heatmaps for ``disgust'' highlight the nose and mouth regions. We hypothesize that the naturally more pronounced wrinkles and folds in these areas in elderly faces resemble the features used by the baseline model to identify disgust in younger subjects. 

        In summary, our analysis confirms that facial expression recognition models trained in a standard supervised fashion suffer from significant age-related biases. The heatmap analysis suggests that these errors are also related to how the model attends to different facial regions depending on age. Altogether, these findings highlight the importance of incorporating age-awareness into model training, and motivate the exploration of targeted bias mitigation strategies, as we discuss next.

    \subsection{RQ2: Can FER models be improved for specific age groups through bias-mitigation techniques?}

        Many previous works mitigate age bias in deep learning by modifying the model architecture~\cite{wu2015enhanced, yang2018joint, huang2024facial}. However, model bias often stems from training data (e.g. representation bias \cite{suresh2021framework, mehrabi2021survey}) rather than from the architecture itself. Therefore, we argue that bias can be addressed without significant model modifications. In this work, we investigated three model-agnostic strategies: (i) Age-weighted Loss, (ii) Multi-task Learning, and (iii) Multi-modal Input. Experiments showed that all three improved performance for the most disadvantaged age group---the elderly---while maintaining overall performance for other groups, contributing to a fairer facial expression recognition across age groups.

        The main improvement for the elderly was for the ``neutral'' expression, previously confused with ``anger'' and ``sadness'' by the baseline. All three strategies significantly reduced this confusion. Additionally, the misclassification of ``sadness'' as ``disgust'' experienced a considerable reduction. In contrast, none of the methods alleviated the confusion between ``fear'' and ``surprise'' in children. Nevertheless, the most important outcome was the reduced performance gap across age groups. For example, the performance difference between adults and elderly dropped from .0917 for ConvNeXt baseline to .0380 for ConvNeXt Age-weighted Loss.

        An interesting outcome was the shift in saliency heatmaps for the improved models. Introducing age-related features accentuated differences in salient regions across age groups. The elderly---who benefited the most in performance---also exhibited the most heatmap changes, suggesting that adapting the attended regions was key to their improved performance. For instance, increased importance of the eyebrows and edges of the mouth for ``sadness'' may have helped reduce misclassifications with ``neutral'' and  ``sadness'' in the elderly. Children and adults, by contrast, showed fewer such changes, consistent with their smaller confusion issues.

        Age-weighted Loss proved powerful in rebalancing age group representation during training, particularly when combined with class-balancing. Furthermore, per-age-group weights were integrated successfully into the other two approaches, confirming that addressing training data imbalance is key to improving model learning for underrepresented groups. The two other approaches, Multi-task Learning and Multi-modal Input, aimed to inject age-related information into the expression classification process. In Multi-task Learning, the model was trained to simultaneously estimate age, forcing the learned feature vector to encode age-relevant features. In Multi-modal Input, age was directly concatenated to the feature vector before classification. Both strategies led to improved performance for the elderly, albeit with different expression-specific outcomes.
        
        Multi-task Learning has previously been used to mitigate age bias, e.g., by Yang et al.~\cite{yang2018joint}, who used a dual-branch architecture (ScatNet for age, ConvNet for expressions) and evaluated performance on the same dataset (FACES and Lifespan) used for training. We argue that such complexity is unnecessary, since age estimation is a secondary task only used to guide the model's attention. A simple parallel fully connected layer for age estimation is sufficient, offering the advantage of architectural simplicity and wide applicability. Other work, such as by Liu et al.~\cite{liu2018exploring} and Alvi et al.~\cite{alvi2018turning}, applied multi-task learning for bias mitigation by \textit{removing} features associated with demographic attributes like age, gender, and ethnicity. While this may be appropriate in domains where fairness requires excluding such factors (e.g., hiring, insurance), it reduces model awareness of age-related variations, which may harm performance. For FER, where the goal is equal performance across groups, we argue that the model should instead learn to identify and \textit{compensate} for such variations.

        Attribute-aware methods, where age is provided as a model input, have also been used~\cite{yang2018joint}. In our Multi-modal Input strategy (see Section~\ref{sec:multi-modal-input}), we adapted this idea by down-scaling the feature vector and concatenating the age value. Although effective, this approach has the obvious drawback of requiring age labels at inference time. By contrast, Multi-task Learning is self-sufficient, since the model learns to estimate age internally. Moreover, including age as a single input feature may be too simplistic, given that aging involves complex facial changes. Multi-task Learning allows the model to learn these nuances directly.        

    \subsection{Additional considerations}


        We chose AffectNet as the training dataset based on prior work showing its size and strong generalization to other FER datasets~\cite{gaya-morey2025evaluating}. However, because AffectNet lacks age labels, we estimated them automatically (see Section~\ref{sec:datasets}). This raises questions about label quality and its impact on performance.

        Despite relying on automated age labels, our models exhibited reduced age bias, indicating that these labels were sufficiently accurate for the purposes of training and bias analysis. While we do not claim that AI-generated labels are equivalent in quality to manual annotations, they offer a scalable alternative for enriching datasets that lack demographic metadata---particularly valuable in bias diagnosis and mitigation, as demonstrated in this work. Notably, training on AffectNet allowed our baseline MobileNetV3, Swin Transformer and ConvNeXt models to achieve approximately .7329, .7535 and .7528 accuracy, respectively, on the elderly subset of the FACES dataset. This significantly outperforms the best results reported by Huang et al.~\cite{huang2024facial}, whose ResNet-18 models trained on RAF-DB achieved .5789 (baseline) and .6784 (with their proposed AEFL method) on the same subset. While architectural differences partly explain the performance gap~\cite{ResNet, MobileNetV3}, we attribute the improvement primarily to AffectNet's greater diversity and generalization capacity~\cite{gaya-morey2025evaluating}.

    \section{Future work}

        The success of our mitigation strategies---and their model- and task-agnostic nature---suggests they could be applied in other domains. For instance, Age-weighted Loss, Multi-task Learning, and Multi-modal Input may prove useful in facial recognition, with applications ranging from authentication to human-computer interaction. These strategies may also mitigate other forms of bias, such as gender or ethnicity.

        One critical limitation is the lack of datasets with elderly individuals. While we accessed several datasets for children and adults (see Section~\ref{sec:datasets}), only FACES included elderly subjects. We advocate for the creation and release of more representative datasets for older demographics. Moreover, we encourage dataset curators to include demographic attributes like age, gender, and ethnicity, which are often missing in large-scale FER datasets.

        While bias was substantially reduced for the elderly, it occasionally came at the cost of lower performance for children and adults. Future work should focus on mitigating bias while retaining performance in majority groups. Additionally, we aim to further investigate the persistent confusion between ``fear'' and ``surprise'' in children---the only major bias that remained unresolved.

\section{Conclusion}

    In this work, we studied the presence of age-related bias in facial expression recognition models, with a particular focus on the elderly population. Our analysis, grounded on per-age-group performance, confusion matrices, and saliency heatmaps, revealed systematic errors that disproportionately affected elderly individuals---most notably the confusion of ``neutral'' with ``sadness'' and ``anger'', and between ``sadness'' and ``disgust''. In contrast, children showed a distinct bias pattern, primarily characterized by confusion between ``fear'' and ``surprise''.

    To address these biases, we proposed and evaluated three mitigation strategies: Age-weighted Loss, Multi-task Learning, and Multi-modal Input. All three approaches resulted in notable performance improvements for elderly individuals, especially for expressions previously most affected by bias. Moreover, the inclusion of age-related information---either explicitly or implicitly---helped the model to adjust its focus on more relevant facial regions for different age groups, as revealed by the heatmap analysis.
    
    Our findings demonstrate that it is possible to mitigate age-related bias without modifying model architecture substantially. Moreover, they suggest that even approximate demographic labels, such as automatically estimated age, can be effective for bias-aware training, enabling the use of large datasets like AffectNet that lack manual demographic annotations.
    
    Future work will focus on further reducing expression confusion in children---particularly the unresolved bias between fear and surprise---and on developing techniques that improve fairness without compromising performance in majority groups. We also emphasize the need for more balanced datasets with adequate representation of elderly individuals, as well as enriched demographic annotations to better assess and mitigate bias in FER systems.

\printcredits



\section*{Acknowledgements}

    This work is part of the Project PID2023-149079OB-I00 (EXPLAINME) funded by MICIU/AEI/10.13039/ 501100011033/ and ERDF, EU and of Project PID2022-136779OB-C32 (PLEISAR) funded by MICIU/ AEI /10.13039/501100011033/ and FEDER, EU. F. X. Gaya-Morey was supported by an FPU scholarship from the Ministry of European Funds, University and Culture of the Government of the Balearic Islands.

\section*{Declaration of competing interest}

    The authors declare that they have no known competing financial interests or personal relationships that could have appeared to influence the work reported in this paper.

\section*{Declaration of generative AI and AI-assisted technologies in the writing process}

    During the preparation of this work the authors used ChatGPT in order to improve the readability and language of the manuscript. After using this tool, the authors reviewed and edited the content as needed and take full responsibility for the content of the published article.

\appendix

\section{Additional results}\label{secA1}

    \subsection{Baseline detailed results}

\begin{figure}[h]
                \centering
                \includegraphics[width=\textwidth]{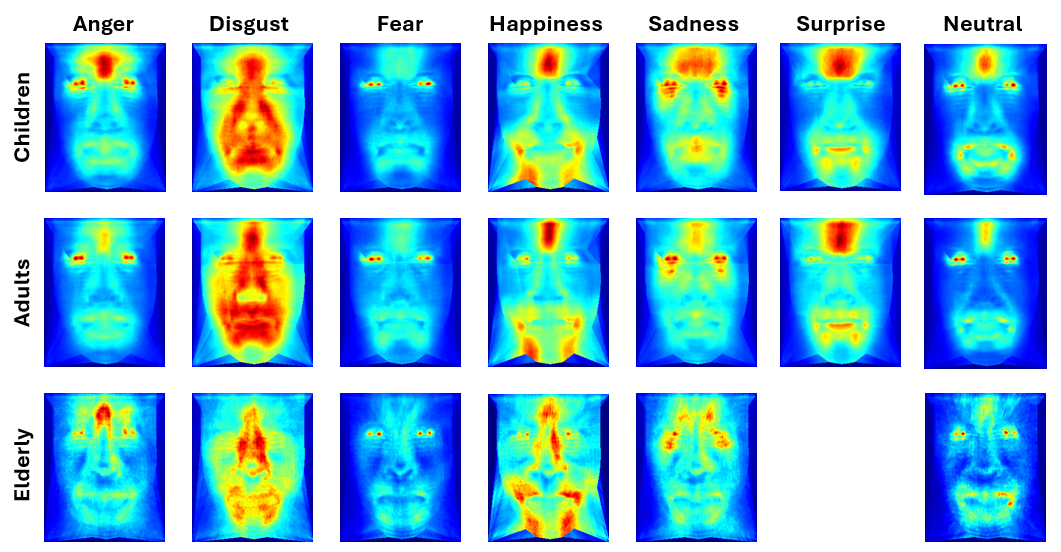}
                \caption{Explanation heatmaps by age group and expression for MobileNetV3 baseline approach.}
                \label{fig:heatmaps-mob-base}
        \end{figure}

\begin{figure}[h]
                \centering
                \includegraphics[width=\textwidth]{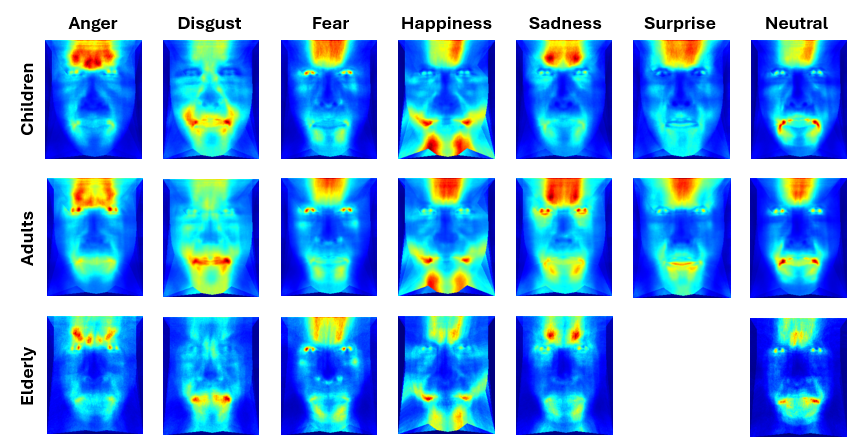}
                \caption{Explanation heatmaps by age group and expression for Swin Transformer baseline approach.}
                \label{fig:heatmaps-swin-base}
        \end{figure}

        As previously discussed in Section~\ref{sec:age-bias-exploration}, we provide a comparative analysis of the explanation heatmaps generated by the ConvNeXt-Base, MobileNetV3 and Swin Transformer baseline approaches, shown in Figure~\ref{fig:heatmaps-base}, Figure~\ref{fig:heatmaps-mob-base} and Figure~\ref{fig:heatmaps-swin-base} respectively.
        
        In general, both the ConvNeXt and Swin Transformer models show a notable consistency in identifying salient regions across expressions, whereas MobileNetV3 exhibits a more dispersed and less focused attention pattern, which may be associated with its lower representational capacity. 
    
        A more detailed examination of individual facial expressions reveals important model-specific distinctions. For the expression of ``anger'', all three models agree on the fundamental regions, although subtle differences are observed: MobileNetV3 tends to highlight the eyes in adults, instead of the brow area. Swin Transformer, on the other hand, places greater importance on the forehead, especially in children and adults. Regarding the ``disgust'' expression, MobileNetV3 activates a considerably wider area compared to ConvNeXt, including both the nose and mouth, but without precise focusing. Swin Transformer, for its part, keeps its attention on the mouth region but attenuates focus on the nose and subtly incorporates attention to the forehead, especially within the children group. In the case of ``fear'', discrepancies between models are particularly notable. ConvNeXt strongly highlights the forehead and chin. MobileNetV3, in contrast, significantly reduces its emphasis on these areas and distributes attention more evenly among the eyes, forehead, and mouth, but with less intensity. Swin Transformer, while closer to ConvNeXt, also minimizes attention on the chin and assigns less relevance to the mouth region. Regarding ``happiness'' expression, all three models show high agreement, although MobileNetV3 introduces a significant difference in the elderly group by attributing greater importance to the nasal region. In the case of ``surprise'' expression, both Swin Transformer and MobileNetV3 align with ConvNeXt focusing their attention on the same salient regions. However, consistent with earlier observations, MobileNetV3’s heatmaps remain more spatially dispersed and less sharply focused. Finally, heatmaps for ``sadness'' and ``neutral'' expressions are broadly comparable between all three models, exhibiting a high degree of overlap in their salient regions.
        
    \subsection{Improvement strategies detailed results}

\begin{figure}[h]
                \centering
                \includegraphics[width=\textwidth]{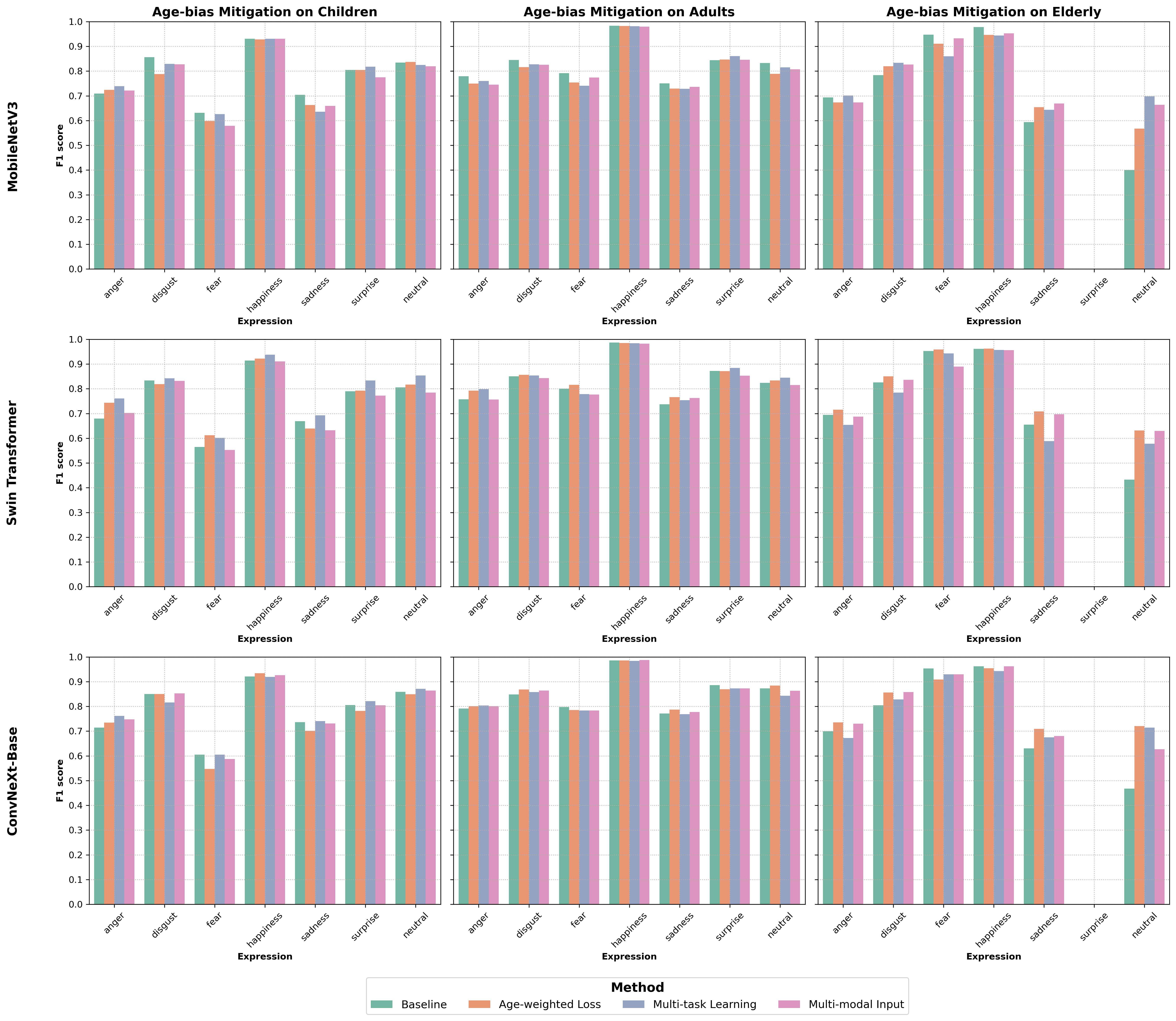}
                \caption{Per-class performance of the different methods addressing age bias. From left to right: children, adults and elderly. From up to bottom: MobileNetV3, Swin Transformer and ConvNeXt-Base.}
                \label{fig:bars-methods}
        \end{figure}

\begin{figure}[h]
                \centering
                \includegraphics[width=\textwidth]{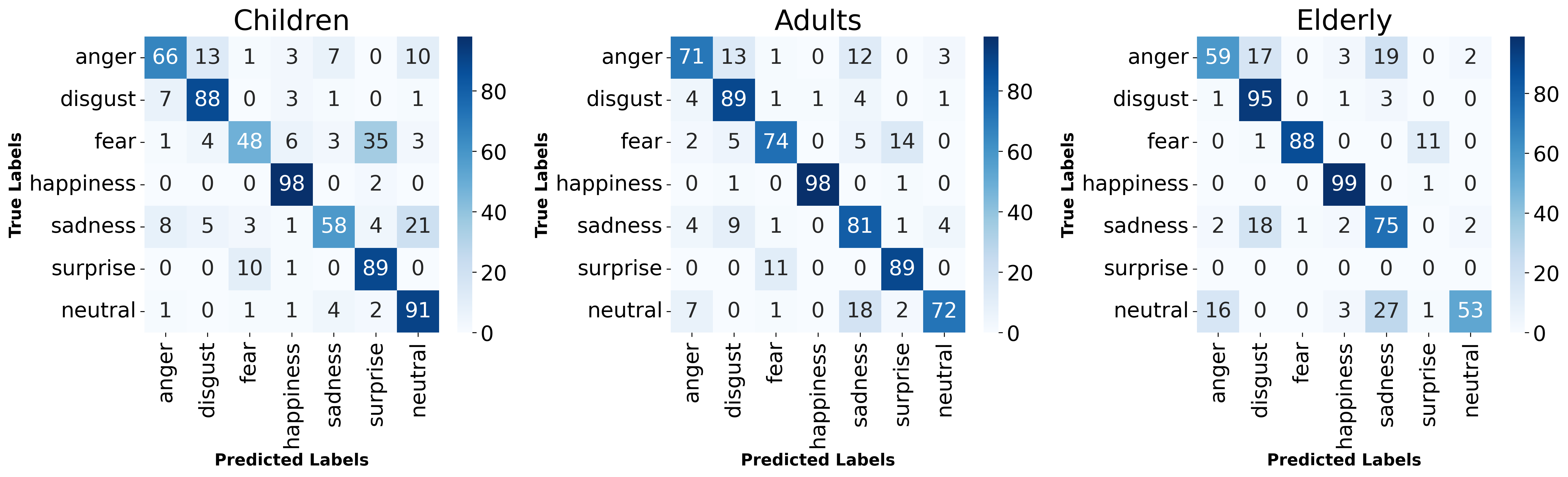}
                \caption{Normalized confusion matrices by age group for the MobileNetV3 multi-modal approach. From left to right: children, adults and elderly.}
                \label{fig:cms-mob-multimod}
        \end{figure}

\begin{figure}[h]
                \centering
                \includegraphics[width=\textwidth]{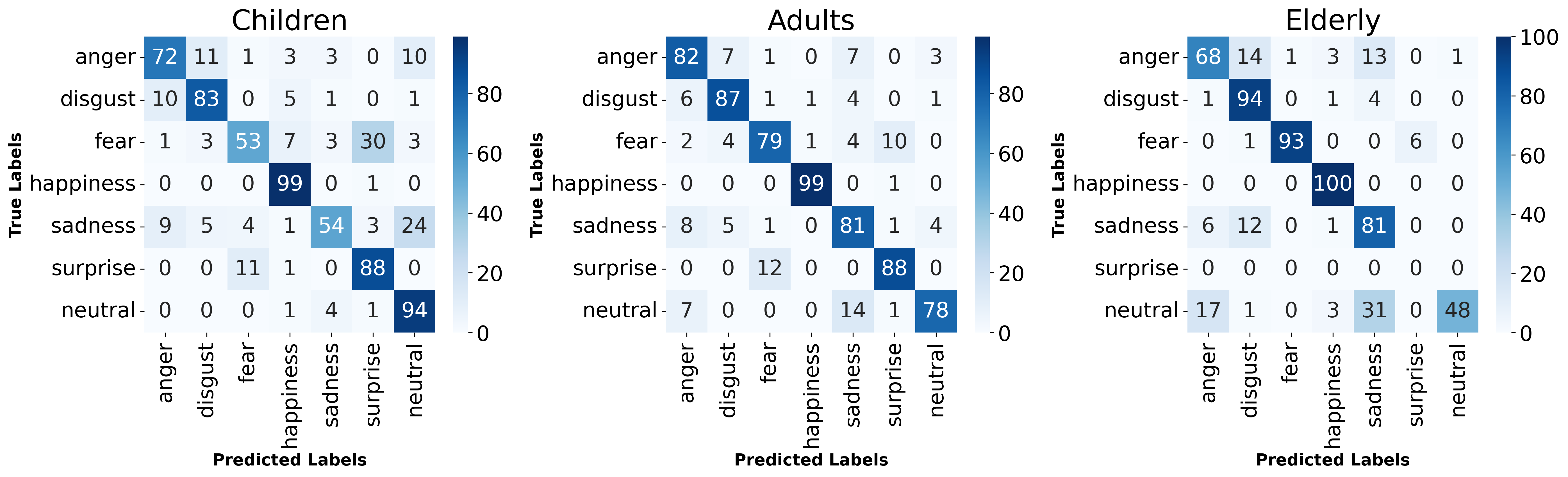}
                \caption{Normalized confusion matrices by age group for the Swin Transformer age-weighted loss approach. From left to right: children, adults and elderly.}
                \label{fig:cms-swin-age-weighted}
        \end{figure}

        Figure~\ref{fig:bars-methods} displays F1 scores by expression and age-group, evaluated for each of the three architectures considered---MobileNetV3, Swin Transformer and ConvNeXt-Base---for each of the three methodological strategies described in Section~\ref{sec:approaches}---Age-weighted Loss, Multi-Task Learning and Multi-modal Input. Baseline performance is also included to facilitate direct comparison across the different approaches. Additionally, the confusion matrices by age group for the best-performing strategies of MobileNetV3 and Swin Transformer, which correspond to Multi-modal Input approach and Age-weighted Loss strategy respectively, are shown in Figures~\ref{fig:cms-mob-multimod} and \ref{fig:cms-swin-age-weighted}. 
        
        Across all three networks studied we observe similar behavioral patterns regardless of their underlying architecture. For the elderly group, similarly to ConvNeXt, the confusion matrices for the improved models of Swin Transformer and MobileNetV3 demonstrate a marked reduction in the misclassification of the ``neutral'' expression as ``anger'' and ``sadness'', as well as a reduction in the misclassification of ``sadness'' as “disgust''. This improvement translates into a substantial increase in true positive rates for both the ``neutral'' and ``sadness'' classes when compared to their respective baseline models. In children, the enhanced models demonstrate an improved prediction of ``neutral'' expressions. However, this enhancement comes at the expense of a decrease in ``sadness'' prediction, primarily due to increased confusion between ``sadness'' and ``neutral'' expressions. Notably, this particular interaction does not manifest in the adult age group. For the middle-age group, no significant differences in performance were observed between the baseline and the enhanced models. Regardless of the age group, we observe that an increase in true positives for ``fear'' is consistently associated with a decrease in confusion with ``surprise'' and vice versa.
        
        Some subtle differences are noticed between the three architectures. While ConvNeXt exhibited a decline in performance for the ``fear'', ``disgust'' and ``anger'' classes in elderly individuals under the improved modality compared to the baseline, this degradation is not observed in the Swin Transformer. On the contrary, Swin maintains or slightly improves performance in these classes. In contrast, MobileNetV3 reflects a performance reduction similar to ConvNeXt, with the most notable decline occurring in the ``anger'' class.
    
\begin{figure}[h]
                \centering
                \includegraphics[width=\textwidth]{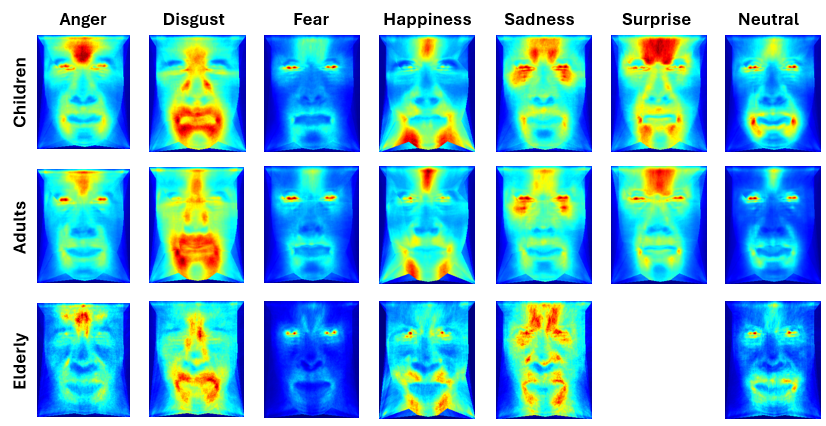}
                \caption{Explanation heatmaps by age group and expression for MobileNetV3 Multi-modal Input approach.}
                \label{fig:heatmaps-mob-multimod}
        \end{figure}

\begin{figure}[h]
                \centering
                \includegraphics[width=\textwidth]{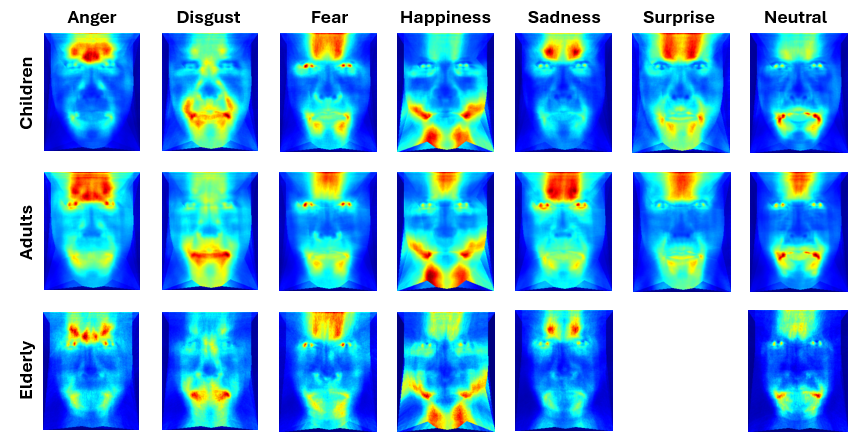}
                \caption{Explanation heatmaps by age group and expression for Swin Transformer age-weighted loss approach.}
                \label{fig:heatmaps-swin-age-weighted}
        \end{figure}

        Figures~\ref{fig:heatmaps-mob-multimod} and \ref{fig:heatmaps-swin-age-weighted} display the corresponding explanation heatmaps for the MobileNetV3 Multi-modal Input model and the Swin Transformer Age-weighted Loss model.
    
        For ``anger'' expression, Swin Transformer displays similar patterns for both baseline and improved heatmaps, although the enhanced version shows slightly higher intensity on the forehead and between the eyebrows. The enhanced MobileNetV3 model also highlights the forehead region more prominently, though this effect is mainly observed in adults. In the case of ``disgust'', saliency is largely consistent, with the improved Swin Transformer exhibiting a marginally increased intensity on the upper nasal region. The enhanced MobileNetV3, for its part, exhibits more focused attention than the baseline on the nasal and oral regions, reducing involvement of surrounding areas, particularly in children and adults. Regarding ``fear'', the improved Swin Transformer model shows increased saliency on the forehead in the elderly group, bringing its pattern more in line with that observed in children and adult age groups. No notable differences were observed for the baseline and improved MobileNetV3 heatmaps for this expression. For ``happiness'', the improved Swin Transformer model slightly reduces the importance attributed to the forehead region. The enhanced MobileNetV3 model shows decreased saliency on the nose in the elderly group, aligning the pattern more closely with that of children and adults. The "sadness" expression shows little change in the Swin Transformer heatmaps. In contrast, the improved MobileNetV3 model increases attention to the mouth area, as well as to the eyes and forehead in the elderly group. In the case of ``surprise'' among children, both improved Swin Transformer and MobileNetV3 show a slight increase in attention around the mouth and chin, especially in children. Finally, no significant differences are observed between baseline and improved models in the ``neutral'' expression for either Swin Transformer or MobileNetV3.
    
\FloatBarrier
    
\section*{Data availability}
All code is made publicly available at \url{https://github.com/Xavi3398/FER-for-elderly}.

\bibliographystyle{cas-model2-names}

\bibliography{bibliography}



\end{document}